\documentclass[8pt]{article}
\usepackage{graphicx}
\usepackage{pgfplots}
\usepackage{newlfont}
\usepackage{tikz}
\usepackage[colorlinks,linkcolor=blue,citecolor=blue,
anchorcolor=blue,bookmarksopen,pdfpagetransition={Wipe}]{hyperref}
\usepackage{amsmath,amsthm,amssymb}
\usepackage{subfigure}
\usepackage{graphicx}
\usepackage{amsmath, amsfonts}
\usepackage{dsfont}
\usepackage{booktabs}
\usepackage{fourier}
\usepackage{array}
\usepackage{makecell}
\usepackage{algorithm}
\usepackage{algpseudocode}
\usepackage{lscape}
\usepackage{threeparttable}
\usepackage{hyperref}
\hypersetup{
	colorlinks=true,
	linkcolor=blue,
	urlcolor=blue
}
\usepackage[T1]{fontenc}
\usepackage[tableposition=top]{caption}
\usepackage{tabularx}

\numberwithin{equation}{section}
\begin{document}
%--------------------------title-----------------Corresponding author:-------
\title{\textbf Semi-Blind Image Deblurring Based on Framelet Prior }
\author{M. Zarebnia$^a$\footnote{Corresponding author:\href{mailto:zarebnia@yahoo.com}{zarebnia@yahoo.com}, \href{zarebnia@uma.ac.ir}{zarebnia@uma.ac.ir}}\
	, R. Parvaz$^a$\footnote{\href{mailto:reza.parvaz@yahoo.com}{reza.parvaz@yahoo.com},
		 \href{rparvaz@uma.ac.ir}{rparvaz@uma.ac.ir} }
}
\date{}
\maketitle
\begin{center}
Department of Mathematics, University of Mohaghegh Ardabili,
56199-11367 Ardabil, Iran.\\
\end{center}
%------------------------------------abstract--------------------------------
\begin{abstract}
\noindent
The problem of image blurring is one of the most studied topics in the field of image processing.
Image blurring is caused by various factors such as hand or camera shake.
To restore the blurred image, it is necessary to know information about the point spread 
function (PSF). And because in the most cases it is not possible to accurately calculate 
the PSF, we are dealing with an approximate kernel.
In this paper, the semi-blind image deblurring problem are studied.
Due to the fact that the model of the deblurring problems is an ill-conditioned problem,
it is not possible to solve this problem directly.
One of the most efficient ways to solve this problem is to use the total variation (TV) method.
In the proposed algorithm,
by using the framelet transform and fractional calculations, the TV method is improved.
The proposed method is used on different types of images and is 
compared with existing methods with different types of tests.
\end{abstract}
%------------------------------------------Keywords---------------------------
\vskip.3cm
\section{Introduction}
\begin{figure}[H]
	\centering
	\makebox[0pt]{
		\includegraphics[width=1\textwidth]{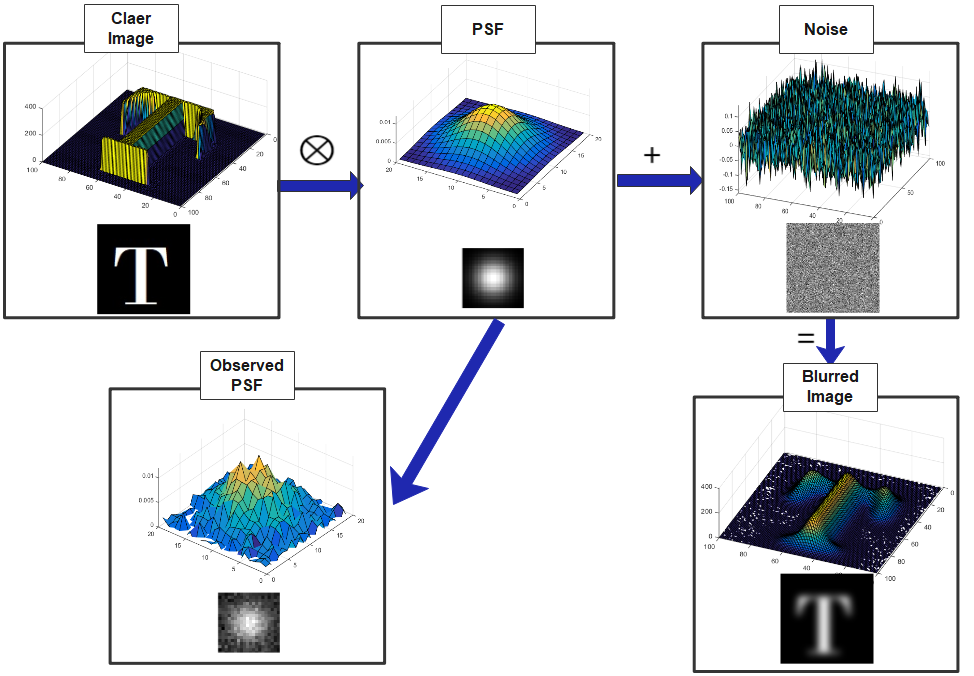}}
	\vspace{-.05cm}
	\caption{Process of image blurring.}
	\label{fignn01}
\end{figure}

Digital image processing is one of the widely used branches of computer science and mathematics.
In this branch of science, various topics such as the image deblurring, object detection and 
image enhancement  are studied.
In this paper, image deblurring is studied.
When taking an image, various factors cause the image to blur, including hand shake or camera shake. 
In addition, the images taken from the sky by the telescope and also the images taken by 
the microscope and medical equipment can be studied in this category of issues. 
Therefore, the study of this topic has many uses in various sciences such as astronomy and medicine.
Assuming that $X$ and $Y$ in $\mathbb{R}^{n\times m}$ indicate the clear and blurred images, respectively. 
Then the model used for the blurred image process is as follows
\begin{align*}
Y=k \otimes X+N.
\end{align*}
In this model, $\otimes$ shows the two dimensional convolution operator, and
$k \in \mathbb{R}^{r\times s}$ represents blur kernel that obtained by point spread function (PSF). 
This function is formulated based on the physical process that causes the blurred image.
For example, the Moffat function is used in astronomical telescope.
The last parameter that appears in this model is $N$.
The effect of noise on the blurred image can be linear or non-linear and can also have different sources.
Among these noises, we can refer to the poisson and gaussian noises.
For a better understanding of this model, the details of the blurring process are given for an example in Figure
\ref{fignn01}.
The method of solving this problem is different depending on the type of noise \cite{a1,a2}.
Also, The image deblurring model can be rewritten as a system of equations as below
\begin{align*}
y=Kx+n,
\end{align*}
where $y,x \in \mathbb{R}^{nm\times 1}$ represent the pixels of the blurred and clear images in vector form, respectively.
$n$ is used for noise.
 Also, $K \in \mathbb{R}^{nm \times nm}$ 
represents the matrix that is obtained according to the PSF and boundary conditions.
For example, if the PSF is nonseparable  and the boundary conditions are considered to be zero, 
then this matrix is the block Toeplitz with Toeplitz blocks (BTTB). The reader can find a full 
discussion of the structure of this matrix in \cite{a3}.
In addition to the ill-conditioned of this problem, the large size of the equations of this system 
also makes it very difficult to directly solve this problem. One of the most effective ways to 
reduce the amount of calculations and increase the speed to solve this problem is to use
the total variation (TV) method. This method is used in \cite{a4} and the effectiveness of this 
method attracted the opinion of researchers. In the following years, TV used in various papers such as \cite{a5,a6}.
To improve the efficiency of this method, various penalty terms and norms have been studied 
in relation to the TV method,
for example \cite{a8,a7}.\\
In this paper, we consider a special case of this type of problem, which is known as the 
semi-blind deblurring problem. 
In this type of problem there is information from the blur kernel along with the error.
The mathematical model of this type of problem is written as follows
\begin{align*}
Y=(k_0+e)\otimes X+N,
\end{align*}
where $k_0$ and $e \in \mathbb{R}^{r\times s}$ denote the observed PSF and error, respectively.
In this model, the value of $y$ and $k_0$ is known and the aim is to approximate $e$ and $x$.
Also, This model can be rewritten as a system of equations as below
\begin{align}\label{eq1}
	y=(K_0+E)x+n,
\end{align}
where $K_0,E \in \mathbb{R}^{nm\times nm}$ according to the boundary conditions and structure 
of $k$ and $e_0$ are obtained.
This problem has been studied in various papers such as \cite{1,2}.  
With the recent development of mathematical science, new tools have been introduced for image processing.
 Among these tools, framelet transform and fractional calculations deserve mention.
These tools have been used in various algorithms to solve the problem of image deblurring, 
among which we can refer to \cite{a9,a10}.
The framelet transform is known to enhance the restored image due to the sparsity of 
representations in image domains. Additionally, when
fractional derivative combined with the total variation model, the restored image 
exhibits distinct and sharp edges.
In the proposed algorithm, these tools are used to improve the total variation method. 
After presenting the proposed model, the alternating direction method of multipliers has 
been used to solve the proposed model, and the results of the method show the improvement of the algorithm.\\

The output of the paper is organized as follows: In Section \ref{sec2}, 
introductory discussions about framelet transform and fractional derivatives are presented 
and then by using these tools a model is introduced for semi-blind image deblurring. 
In Section \ref{sec3}, the numerical algorithm based on the alternating direction method of multipliers is
introduced for the proposed model. Numerical results and analysis related to the proposed algorithm
are given in Section \ref{sec4}. The summary of the paper is presented in Section \ref{sec5}.

\section{Preliminaries and proposed model} \label{sec2}
In this section, in the first subsection, the preliminary topics used in 
this paper is presented, and then the proposed model for semi-blind image deblurring is introduced.

\subsection{Framelet transform and  discrete fractional-order gradient}
\begin{figure}[H]
	\centering
	\makebox[0pt]{
		\includegraphics[width=0.5\textwidth]{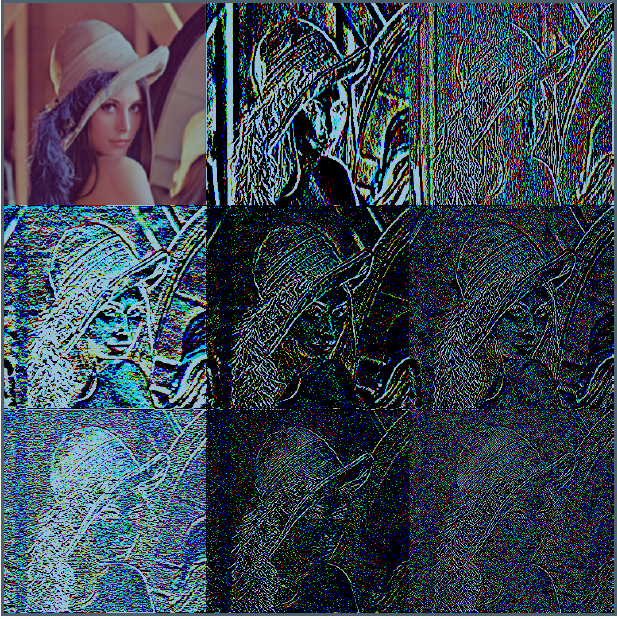}}
	\vspace{-.05cm}
	\caption{Example of framelet transform.}
	\label{figmm01}
\end{figure}

In this subsection, the concepts of framelet transform are briefly studied. The reader can 
find more information about this concept 
in \cite{a11}.
Let $\phi=\{\varphi_\mu\}_{\mu=1}^{r}\subset L^2(\mathbb{R}^d)$, 
if there are two constants $A$ and $B$, so that for the affine system
\begin{align*}
\psi_{\mu,j,k}:=2^{\frac{jd}{2}}\varphi_{\mu}(2^jt-k),~~j\in \mathbb{Z},k\in\mathbb{Z}^d,\mu=1,\cdots,r,
\end{align*}
the following relations is satisfied
\begin{align*}
A \|f\|^2 \leq \sum_{\mu,j,k}|\langle f,\psi_{\mu,j,k} \rangle|^2 \leq B \| f\|^2,~~~~\forall f\in L^2(\mathbb{R}^d),
\end{align*}
where $\langle \cdot,\cdot \rangle$ and $\|\cdot \|$ are shown the inner product and norm 
in $L^2(\mathbb{R})$, respectively,
then $\phi$ is called framelet. When $A=B$, $\phi$
is named tight framelet and if $A=B=1$, $\phi$ is named Parseval frame. Also, when $\|\varphi_\mu\|=1$,
Parseval frame is called wavelet. By using the synthesis and analysis operators \cite{a11}, the 
frame operator is considered as
$S_{\phi}:L^2(\mathbb{R}^d)\rightarrow L^2(\mathbb{R}^d),~S_{\phi}f=\sum_{\mu} \langle f,
\varphi_{\mu}\rangle \varphi_{\mu}$, where 
$f \in L^2(\mathbb{R}^d)$. By using this operator, we can write
\begin{align*}
f=\sum_{\mu} \langle f, S^{-1}_{\phi} \varphi_{\mu} \rangle \varphi_{\mu}
=\sum_{\mu} \langle f, \varphi_{\mu} \rangle S^{-1}_{\phi} \varphi_{\mu},
~~~~\forall f\in L^2(\mathbb{R}^d).
\end{align*}
Therefore, based on the above relation, an image can be expanded by the frame. 
In this paper, Parseval frame that is constructed from B-spline with refinement mask
$h_0=1/4[1,2,1]$ and two framelet masks $h_1=\sqrt{2}/4[1,0,-1]$ 
and $h_2=1/4[-1,2,-1]$ used for image transform.
In the rest of this paper, the matrix of framelet transform is shown with $W$.
Also, an example for this type of transform is given in Figure \ref{figmm01}.
\\

Another concept used in the proposed model is fractional derivative.   
There are various definitions for fractional derivatives, but in this
paper, we focus on the G-L fractional-order derivatives specifically
designed for functions with a discrete structure \cite{a12}.  
Consider $\phi^{\alpha}_i:=(-1)^i\frac{\Gamma(\alpha+1)}{\Gamma(i+1)\Gamma(\alpha+1-i)}$,
the discrete fractional-order gradient of order $\alpha\in \mathbb{R}^+$ 
is defined as $\nabla^{\alpha}X=[D^{\alpha}_hX,D^{\alpha}_vX]^T$ where
horizontal and vertical derivatives are obtained as
\begin{align*}
D^{\alpha}_hX_{i,j}=\sum_{l=0}^{q-1}\phi_l^{\alpha}X_{i-l,j},~~~~D^{\alpha}_vX_{i,j}
=\sum_{l=0}^{q-1}\phi_l^{\alpha}X_{i,j-l}.
\end{align*}
In the above formulas, $q$ represents the  number of neighboring pixels. In this paper, 
this value is selected 
equal to 15 for the numerical results. 

\subsection{Proposed model}
The proposed model to solve the problem \eqref{eq1} is presented as follows
\begin{align}\label{eq2}
(x,E)=\arg \min_{x,E} \frac{1}{2}\|(K_0+E)x-y\|^2+\sum_{i=1}^2 \lambda_i \| \varpi_i x\|
+\frac{\lambda_3}{2}\|E \|^2+\delta_{[0,1]} (x),
\end{align}
where $\varpi_1=W,\varpi_2=\nabla ^{\alpha}$, and  $\| \cdot \|$ denotes the 2-norm 
and $\delta$ is a convex projection operator where defined as
\begin{align*}
	\delta_{[0,1]}(x)=
	\begin{cases}
		0, & \text{if } x \in [0,1],\\
		\infty, & \text{if } x \notin [0,1].
	\end{cases}
\end{align*}
Also, in this minimization problem, $\{\lambda_i\}^3_{i=1}$ are considered as positive parameters.
According to the definition of jointly convex, this model is clearly not a jointly convex model for 
(x,e).
However, as will be seen in the rest of the paper, this model can be divided into two sub-models.
As can be seen in the proposed model, framelet transform and  discrete fractional-order gradient are used.
Combining the framelet transform and fractional derivative with the total
 variation model has been observed to improve the quality of restored images. 
 This is primarily because the framelet transform promotes sparsity in image 
 representations, leading to accurate restoration. Moreover, by using fractional 
 derivative, the restored image obtain sharp edges.  In the next section, 
 an algorithm for the numerical solution of this model is presented.

\section{The solution of the proposed model} \label{sec3}
The problem \eqref{eq2} can be divided into the following two sub-problems.
To solve the problem \eqref{eq2}, the iterative method is used. 
Suppose that at the $k$th step the values $x^k$ and $e^k$ have been calculated, then 
at the  $(k+1)$th step, $x^{k+1}$ can be calculated by solving the following sub-problem.
\begin{align}
x^{k+1}=\arg \min_{x}&\frac{1}{2}\|(K_0+E^k) x-y\|^2+\sum_{i=1}^2 \lambda_i \| \varpi_i x\|\nonumber\\ \label{eq3}
&+\delta_{[0,1]}(x)+\frac{\beta_1}{2}\|x-x^k\|^2.
\end{align}
In this formula, $\beta_1$ is regarded as positive parameter and the last sentence has been 
added to control and minimize any significant alterations.
likewise, by considering  $\beta_2$ as positive parameter, after calculating
 $x^{k+1}$, $e^{k+1}$ is obtained as follows.
\begin{align}\label{eq4}
	E^{k+1}=\arg \min_{E} \frac{1}{2}\|(K_0+E)x^{k+1}-y\|^2
	+\frac{\lambda_3}{2}\|E \|^2+\frac{\beta_2}{2}\| E-E^k\|.
\end{align}

\subsection{Solving subproblems}
In this subsection, the method of solving the subproblems is presented.
In the first step, the alternating direction method of multipliers (ADMM) is used 
for the subprobelem \eqref{eq3}.
By auxiliary variables $\{\eta_i\}_{i=1}^3$, the subprobelem \eqref{eq3} can be written as
\begin{align*}
	&x^{k+1}=\arg \min_{x}\frac{1}{2}\|(K_0+E^k) x-y\|^2+\sum_{i=1}^2\lambda_i \|\eta_i\|
	+\delta_{[0,1]}(\eta_3)+\frac{\beta_1}{2}\|x-x^k\|^2,\\
	& s.t: \eta_1=Wx,~~\eta_2=	\nabla ^{\alpha} x,~~\eta_3=x.
\end{align*}
Then the augmented Lagrangian for this problem is obtained as
\begin{align*}
L(x,\eta_1,\eta_2,\eta_3)&=\frac{1}{2}\|(k_0+E^k) x-y\|^2+\sum_{i=1}^2\lambda_i \|\eta_i\|
+\delta_{[0,1]}(\eta_3)\\
&+\frac{\beta_1}{2}\|x-x^k\|^2+\sum_{i=1}^3 \big(\langle \theta_i,\varpi_i x-\eta_i\rangle
+\frac{\beta_3}{2}\| \varpi_i x-\eta_i\|^2\big),
\end{align*} 
where $\beta_3 >0$ and $\varpi_3=I$ (identity matrix).
Then the extended iterative algorithm for ADMM is obtained as
 \begin{align}
 	&x^{k+1}=\arg \min_{x}\frac{1}{2}\|(K_0+E^k) x-y\|^2+\sum_{i=1}^3\big( \langle \theta_i,\varpi_i x
 	-\eta_i\rangle +\frac{\beta_3}{2}\| \varpi_i x-\eta_i\|^2\big) \nonumber \\\label{eq5}
 	&~~~~~~~~+\frac{\beta_1}{2}\|x-x^k\|^2,\\\label{eq6}
 	&\eta^{k+1}_i=\arg \min_{\eta_i}\lambda_i \|\eta_i\|+\langle \theta^k_i,\varpi_i  x^{k+1}
 	-\eta_i\rangle+\frac{\beta_3}{2}\| \varpi_i x^{k+1}-\eta_i\|^2,~~i=1,2,\\\label{eq7}
 	&\eta^{k+1}_3=\arg \min_{\eta_3}\delta_{[0,1]}(\eta_3)+\langle \theta^k_3, x^{k+1}
 	-\eta_3\rangle+\frac{\beta_3}{2}\|x^{k+1}-\eta_3\|^2,\\\label{eq8}
 	&\theta_i^{k+1}=\theta^{k}_i+\beta_3(\varpi_i x^{k+1}-\eta_i^{k+1}),~~i=1,2,3.
  \end{align}
To solve problem \eqref{eq5}, the periodic boundary condition is considered. Since by 
using this condition, the blur matrix can be calculated by fast Fourier transform \cite{a3}. 
Then the solution of \eqref{eq5}, by considering the periodic condition and using the 
optimal condition is obtained as
follows.
\begin{align} \label{eq9}
x^{k+1}=\mathcal{F}^{-1}\Big[\frac{\mathcal{F}\big[(K_0+E^k)^{\ast}y+\beta_1x^k
	+\sum_{i=1}^{3}\varpi^{\ast}_i(\beta_3\eta_i-\theta_i)
	\big]}{\mathcal{F}\big[(K_0+E^k)^{\ast}(K_0+E^k)+(\beta_1+2\beta_3)I
	+\beta_3(\nabla^{\alpha})^{\ast}\nabla^{\alpha}\big]}\Big],
\end{align}
where $\ast$, $\mathcal{F}$ and $\mathcal{F}^{-1}$ represent the complex conjugacy, fast 
Fourier transform and inverse fast Fourier transform, respectively. The closed form solution 
of \eqref{eq6}, by using the proximal mapping \cite{b1}, can be written as
\begin{align}\label{eq10}
\eta_i^{k+1}=\max\Big\{\| \varpi_i x^{k}+\frac{\theta^{k}_i}{\beta_3}\|-\frac{\lambda_i}{\beta_3},0\Big\}
\frac{\varpi_i x^{k}+\frac{\theta^{k}_i}{\beta_3}}{\| \varpi_i x^{k}+\frac{\theta^{k}_i}{\beta_3}\|},~~i=1,2.
\end{align}
Also, the solution of \eqref{eq7} can be obtained as
\begin{align}\label{eq11}
\eta_3=\max \Big\{0,\min\big\{1,x^{k}+\frac{\theta^k_3}{\beta_3}\big\}\Big\}.
\end{align}
In summary, based on the relations mentioned in the previous part, the algorithm for 
calculating the value of $x^{k+1}$ can be expressed as Algorithm \textcolor{blue}{1}.

\noindent\line(1,0){280}\\
\vspace{-0.3cm}
Algorithm 1.\\
\vspace{-0.3cm}
\noindent\line(1,0){280}\\
\begin{algorithmic}
	\State \textbf{Input:} $y$, $x^k$, $K_0$, $E^k$, $\beta_1$, $\beta_3$,
	 $\lambda_1$, $\lambda_2$, $\{\theta^0_i\}^3_{i=1}$, $j=0$;
	\Repeat 
	\State obtain $x^j$ by using \eqref{eq9};
	\State obtain $\eta^j_i,~i=1,2$ by using \eqref{eq10};
	\State obtain $\eta^j_3$ by using \eqref{eq11};
    \State update $\theta^j_i$ by using \eqref{eq8};
    \State $j=j+1$;
	\Until{Converged;}
	\State \textbf{Output:} Deblurred image $x^{k+1} \leftarrow x^j$.
\end{algorithmic}
\vspace{-0.3cm}
\line(1,0){280}\\

As the last step in solving the sub-problems, using fast Fourier 
transform, the solution of \eqref{eq4} is calculated as follows.
\begin{align}\label{eq12}
	E^{k+1}=\mathcal{F}^{-1}\Big[\frac{\mathcal{F}\big[(y-K_0x^{k+1})
		(x^{k+1})^{\ast}+\beta_2E^k\big]}{\mathcal{F}\big[ x^{k+1}
		(x^{k+1})^{\ast}+(\lambda_3+\beta_2)I\big]}\Big].
\end{align}

Therefore, based on the above discussion, the final algorithm for calculating the clear image 
is expressed as Algorithm \textcolor{blue}{2}.
The proposed method is a convergent method and the convergence of the proposed method can be
 proven with the exact same method as in articles \cite{1} and \cite{2}. In the simulation results section,
 the convergence of the method is studied using the obtained results.

\noindent\line(1,0){280}\\
\vspace{-0.3cm}
Algorithm 2.\\
\vspace{-0.3cm}
\noindent\line(1,0){280}\\
\begin{algorithmic}
	\State \textbf{Input:} The maximum number of iterations (MaxIt),$k=0$,
	\State \hspace{1cm} the tolerance (tol), $\{\lambda_i\}^3_{i=1}$, 
	$\beta_1$, $u^0$ and $E^0$ as start values;
	\Repeat 
	\State obtain $x^k$ by using Algorithm \textcolor{blue}{1};
	\State obtain $E^k$ by solving \eqref{eq12};
	\State $k=k+1$;
	\Until{Error$=\frac{\|u^{k+1}-u^{k}\|}{\| u^{k+1}\|}\leq$ tol or $k\leq$ MaxIt;}
	\State \textbf{Output:} Deblurred image $x\leftarrow x^k$.
\end{algorithmic}
\vspace{-0.3cm}
\line(1,0){280}\\

\section{Simulation results} \label{sec4}
In this section, we study the results of the algorithm described in the previous section 
and demonstrate the efficiency of the algorithm through various tests.
\subsection{Implementation platform and dataset details}
The system used for simulation includes 
Windows 10-64bit and Intel(R) Core(TM)i3-5005U CPU@2.00GHz.
Also, MATLAB 2014b and its internal functions are used in all simulations.
Internal functions \texttt{fspecial} and \texttt{wgn}  are used to generate the PSF and 
white Gaussian noise, respectively.
To calculate the n-by-m noise matrix, the structure of the function \texttt{wgn}  is 
considered as \texttt{N=wgn(n,m,p,$'$dBm$'$)},
and it should be noted that the \texttt{dBm} unit is different from the \texttt{dB} unit.
Additionally, other units such as \texttt{dBW} and \texttt{ohm} can be used for calculations.
Also, the function \texttt{imfilter(-,-,} \texttt{$'$circular$'$)} is used to produce the blurred image.
After calculating the value of $k$ by \texttt{fspecial}, the PSF error matrix $e$ is calculated
as \texttt{e=std*randn} \texttt{(-)}, and then the matrix $k_0$ is considered as $k_0=k-e$.
The images used in this section are from the USC-SIPI images
database \footnote{ \href{http://sipi.usc.edu/database/}{http://sipi.usc.edu/database/} }.
In this section,  the command \texttt{rgb2gray} is used to convert a color image to a gray image.
Also, in cases where the size of the image is different from the size of the source, the command 
\texttt{imresize} is used to reduce the size of the image. In the calculations related to the 
proposed algorithm, the value of tol is considered as $10^{-3}$.
Also, in the numerical experiments, the following ranges are selected for input parameters:
$\lambda_1, \lambda_2, \lambda_3 \in \{10^{-6},10^{-5},10^{-4},10^{-3},10^{3},10^{5}\}$, $\beta_1, 
\beta_2\in\{0.1,1,10\}$,
$\beta_3 \in \{10^{i}; i=-6,\cdots,0\}$
and $\alpha \in\{0.25,0.5,0.75,1,1.5,1.75\}$.
After restoring the image, three quantities are used to evaluate the numerical 
results and compare them with other methods: peak signal-to-noise ratio (PSNR), 
structural similarity (SSIM), and feature similarity (FSIM). More information about 
these values can be found in  \cite{b2}.

\subsection{Numerical experiments}
\begin{figure}[H]
	\centering
	\makebox[0pt]{
		\subfigure[]{\label{fig:gull}\includegraphics[width=0.35\textwidth]{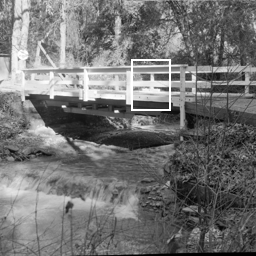}}
		\subfigure[]{\label{fig:tiger}\includegraphics[width=0.35\textwidth]{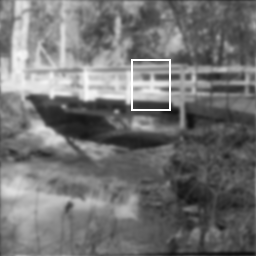}}
	\subfigure[]{\label{fig:mouse}\includegraphics[width=0.35\textwidth]{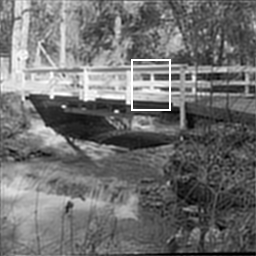}}}\\
			\makebox[0pt]{
		\subfigure[]{\label{fig:mouse}\includegraphics[width=0.18\textwidth]{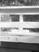}}
			\hspace{1.2cm}
		\subfigure[]{\label{fig:mouse}\includegraphics[width=0.18\textwidth]{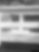}}
		\hspace{1.2cm}
		\subfigure[]{\label{fig:mouse}\includegraphics[width=0.18\textwidth]{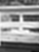}}		
	}
	\vspace{-.2cm}
	\caption{(a,d) Ground-truth image, (b,e) degraded image with $k_1$, (c,f) restored image by proposed algorithm.}
	\label{fig1}
\end{figure}

\begin{figure}[H]
	\centering
	\makebox[0pt]{
		\subfigure[]{\label{fig:gull}\includegraphics[width=0.35\textwidth]{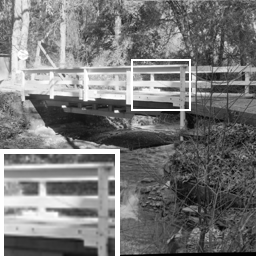}}
		\subfigure[]{\label{fig:tiger}\includegraphics[width=0.35\textwidth]{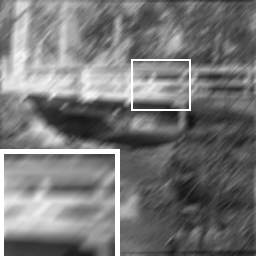}}
}\\
	\makebox[0pt]{
				\subfigure[]{\label{fig:tiger}\includegraphics[width=0.35\textwidth]{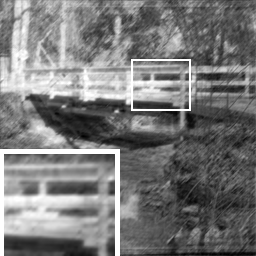}}
		\subfigure[]{\label{fig:mouse}\includegraphics[width=0.35\textwidth]{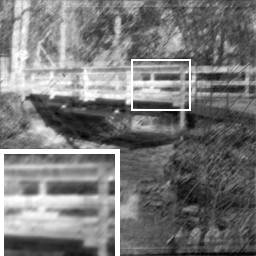}}
		\subfigure[]{\label{fig:mouse}\includegraphics[width=0.35\textwidth]{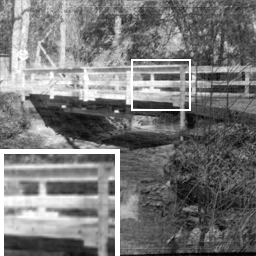}}	
	}
	\vspace{-.2cm}
	\caption{(a) Ground-truth image, (b) degraded image with $k_2$, restored image by: 
		(c) method in \cite{2}, (d) method in \cite{1}, (e) proposed algorithm.}
	\label{fig2}
\end{figure}

\begin{figure}[H]
	\centering
	\makebox[0pt]{
		\subfigure[]{\label{fig:gull}\includegraphics[width=0.45\textwidth]{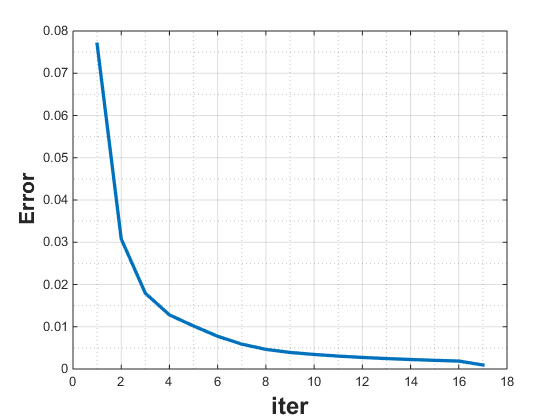}}
		\subfigure[]{\label{fig:tiger}\includegraphics[width=0.45\textwidth]{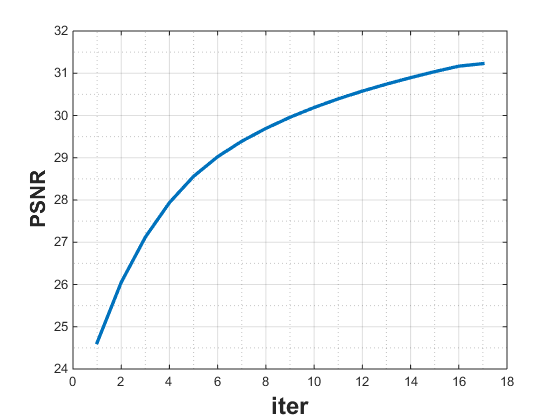}}
	}\\
	\makebox[0pt]{
		\subfigure[]{\label{fig:gull}\includegraphics[width=0.45\textwidth]{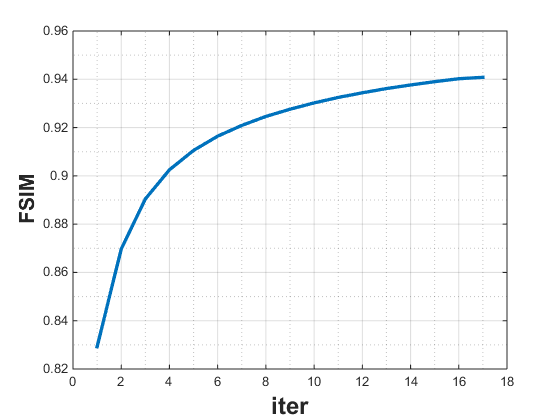}}
		\subfigure[]{\label{fig:tiger}\includegraphics[width=0.45\textwidth]{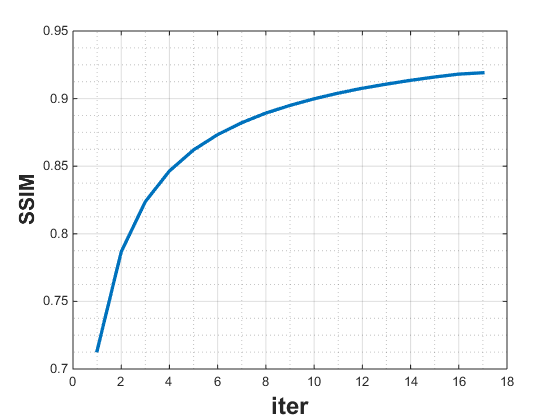}}
	}
	\vspace{-.2cm}
	\caption{Error, PSNR, FSIM and SSIM curves for \texttt{5.2.10} with $k_2$.}
	\label{fig3}
\end{figure}

As first example, \texttt{5.2.10} ($256 \times 256$), 
\texttt{$k_1$=fspecial($'$gaussian$'$,} \texttt{[15 15]} \texttt{,1.5)},
 \texttt{$k_2$=fspecial($'$motion$'$,10,45)} and \texttt{std=0.001} 
 are considered. Also, in this section noise matrix is regarded as 
 \texttt{$N$=wgn(-,-,4,$'$dBm$'$)}. In Table \ref{Tab1}, the results of 
 the proposed method are compared with the results of other methods for two PSFs.  
For another experiment, the results of restored image and its enlargement part 
for PSF $k_1$ are given in Figure \ref{fig1}.
The results show the proper restore of the clear image. 
Also, in Figure \ref{fig2}, the results of the proposed algorithm are compared 
with the methods presented in \cite{1} and \cite{2}. In this figure, a portion 
of the image is enlarged to demonstrate the effect of the method on the image. 
By analyzing these results, it can be seen that the proposed algorithm 
improves the quality of the restored image.
To check the convergence of the proposed algorithm, the figures of error, PSNR, SSIM and FSIM are given in Figure \ref{fig3}. The results of this figure shows the convergence of the method.

\begin{center}
	\begin{table}[H]
		\begin{threeparttable}
			\caption{ Results for \texttt{5.2.10} ($256 \times 256$) with different PSF.}
			\label{Tab1}
			\centering
			\begin{tabular}{cccccccccc}
				\hline
				&\multicolumn{3}{c}{method in \cite{2}}   &\multicolumn{3}{c}{method in \cite{1}  }   &\multicolumn{3}{c}{proposed method}\\
				\cmidrule(l){2-4} \cmidrule(l){5-7} \cmidrule(l){8-10}
				PSF &PSNR      &FSIM   &SSIM  &PSNR         &FSIM&        SSIM      &PSNR      &FSIM     &SSIM\\
				\hline
				$k_1$      &26.1201   &0.9001   &0.8116&26.3603    &0.9118    &0.8316 &27.9004    &0.9208    &0.8501\\
				$k_2$       &26.3309   &0.8792   &0.8059 &26.7180     &0.8809    &0.8062   &31.2280    &0.9408    &0.9191\\
				\hline
			\end{tabular}
		\end{threeparttable}
	\end{table}
\end{center}

%===============================================================================================

\begin{figure}[H]
	\centering
	\makebox[0pt]{
		\subfigure[degraded image]{\label{fig:gull}\includegraphics[width=0.25\textwidth]{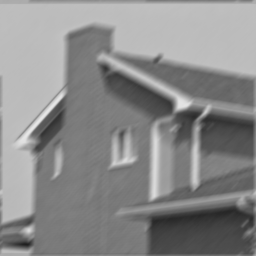}}
		\subfigure[method in \cite{2}]{\label{fig:tiger}\includegraphics[width=0.25\textwidth]{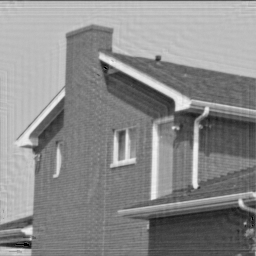}}
				\subfigure[method in \cite{1}]{\label{fig:gull}\includegraphics[width=0.25\textwidth]{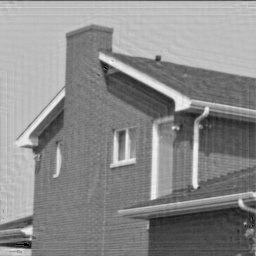}}
		\subfigure[proposed method]{\label{fig:tiger}\includegraphics[width=0.25\textwidth]{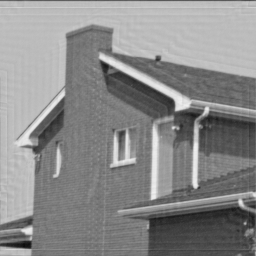}}
	}\\
	\makebox[0pt]{
		\subfigure[degraded image]{\label{fig:gull}\includegraphics[width=0.25\textwidth]{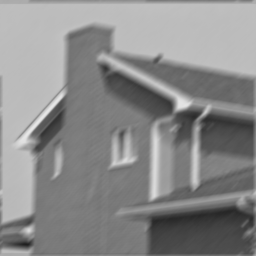}}
\subfigure[method in \cite{2}]{\label{fig:tiger}\includegraphics[width=0.25\textwidth]{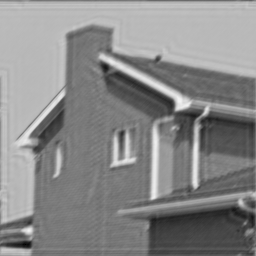}}
\subfigure[method in \cite{1}]{\label{fig:gull}\includegraphics[width=0.25\textwidth]{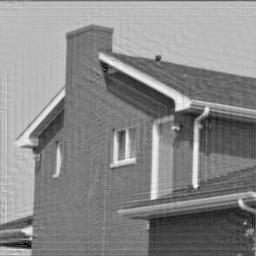}}
\subfigure[proposed method]{\label{fig:tiger}\includegraphics[width=0.25\textwidth]{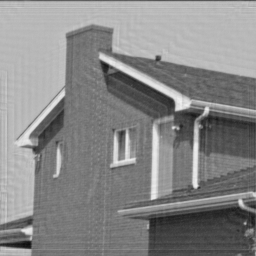}}
	}\\
	\makebox[0pt]{
	\subfigure[degraded image]{\label{fig:gull}\includegraphics[width=0.25\textwidth]{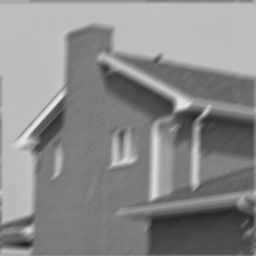}}
	\subfigure[method in \cite{2}]{\label{fig:tiger}\includegraphics[width=0.25\textwidth]{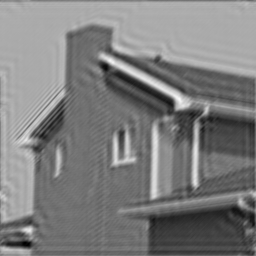}}
	\subfigure[method in \cite{1}]{\label{fig:gull}\includegraphics[width=0.25\textwidth]{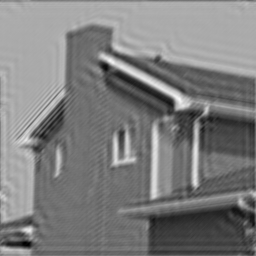}}
	\subfigure[proposed method]{\label{fig:tiger}\includegraphics[width=0.25\textwidth]{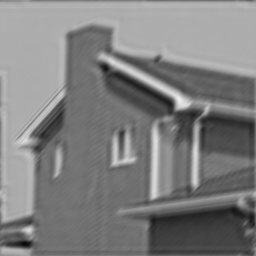}}
}
	\vspace{-.2cm}
	\caption{Visual comparison for \texttt{4.1.05} with: 
		(a-d) \texttt{std}=0.0005, (e-h) \texttt{std}=0.005, (i-l) \texttt{std}=0.05.}
	\label{fig4}
\end{figure}

\begin{figure}[H]
	\centering
	\makebox[0pt]{
		\subfigure[]{\label{fig:gull}\includegraphics[width=0.45\textwidth]{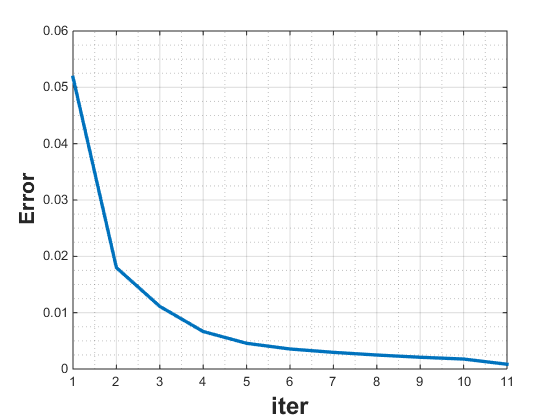}}
		\subfigure[]{\label{fig:tiger}\includegraphics[width=0.45\textwidth]{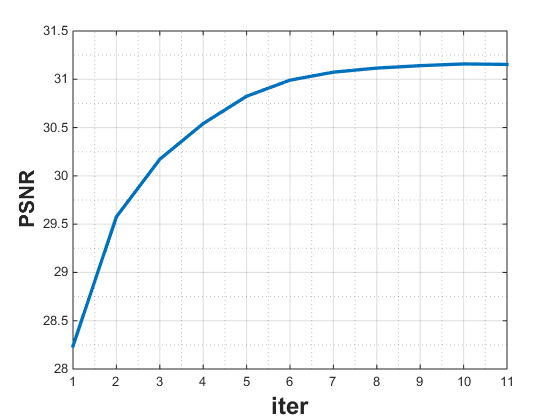}}
	}\\
	\makebox[0pt]{
		\subfigure[]{\label{fig:gull}\includegraphics[width=0.45\textwidth]{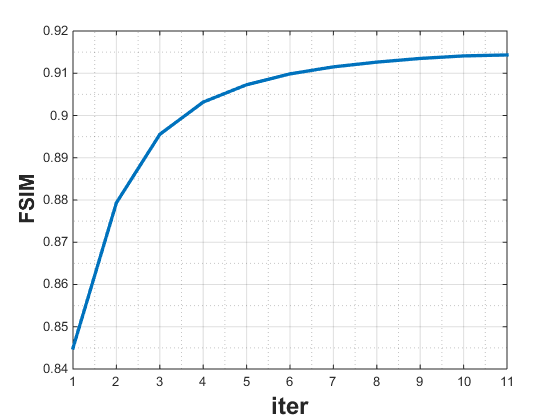}}
		\subfigure[]{\label{fig:tiger}\includegraphics[width=0.45\textwidth]{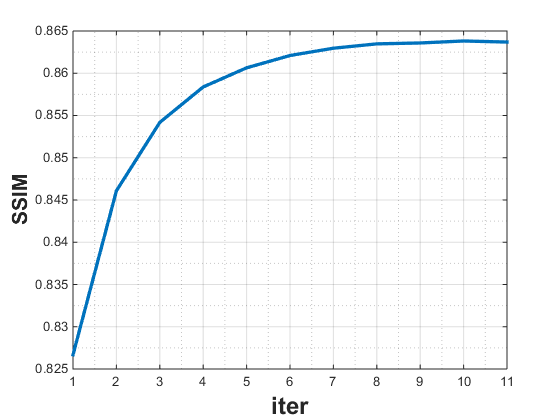}}
	}
	\vspace{-.2cm}
	\caption{Error, PSNR, FSIM and SSIM curves for \texttt{4.1.05} with \texttt{std}=0.005.}
	\label{fig5}
\end{figure}

\noindent In the next example, \texttt{4.1.05} ($256 \times 256$) is 
chosen. For this example, 
$k=$ \texttt{fspecial} \texttt{($'$motion$'$,10,45)} is considered as PSF. 
In Table \ref{Tab2}, the results of various algorithms and the proposed 
algorithm are compared for different values of the \texttt{std}.
Visual comparison with different values of \texttt{std} for proposed method and methods in
\cite{1,2} is given in Figure \ref{fig4}. 
By comparing the results presented for this example, 
it is evident that the proposed method outperforms 
the other methods being compared.
Also, in order to check the convergence of the 
proposed algorithm, the Error, PSNR, FSIM and SSIM curves are drawn in Figure \ref{fig5}.

\begin{center}
	\begin{table}[H]
		\begin{threeparttable}
			\caption{Results for \texttt{4.1.05} ($256 \times 256$) with different values of \texttt{std}.}
			\label{Tab2}
			\centering
			\begin{tabular}{cccccccccc}
				\hline
				&\multicolumn{3}{c}{method in \cite{2}}   &\multicolumn{3}{c}{method in \cite{1}  }   &\multicolumn{3}{c}{proposed method}\\
				\cmidrule(l){2-4} \cmidrule(l){5-7} \cmidrule(l){8-10}
				\texttt{std} &PSNR      &FSIM   &SSIM  &PSNR         &FSIM&        SSIM      &PSNR      &FSIM     &SSIM\\
				\hline
				0.0005     &32.2131   &0.9201   &0.8721  &32.6962    &0.9308    &0.8833 &33.5727    &0.9367    &0.8939\\
				0.005       &30.6838   &0.8983  &0.8212 &30.7978     &0.9021   &0.8392   &31.7551    &0.9254    &0.8764\\
				0.05       &22.5320  &0.7832   &0.7195&22.8697     &0.7969    &0.7254   &23.1754    &0.7979    &0.7674\\
				\hline
			\end{tabular}
		\end{threeparttable}
	\end{table}
\end{center}

\begin{figure}[H]
	\centering
	\makebox[0pt]{
		\subfigure[]{\label{fig:gull}\includegraphics[width=0.35\textwidth]{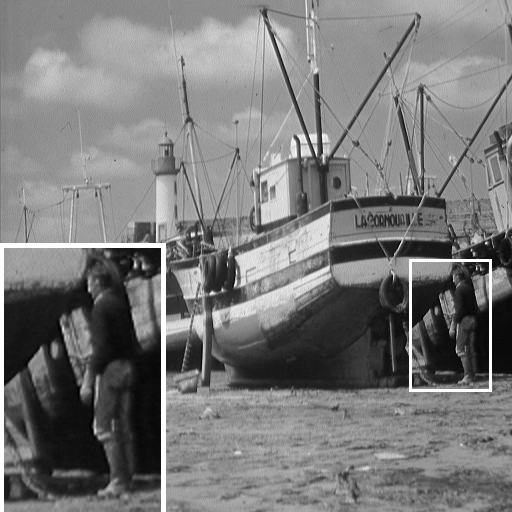}}
		\subfigure[]{\label{fig:tiger}\includegraphics[width=0.35\textwidth]{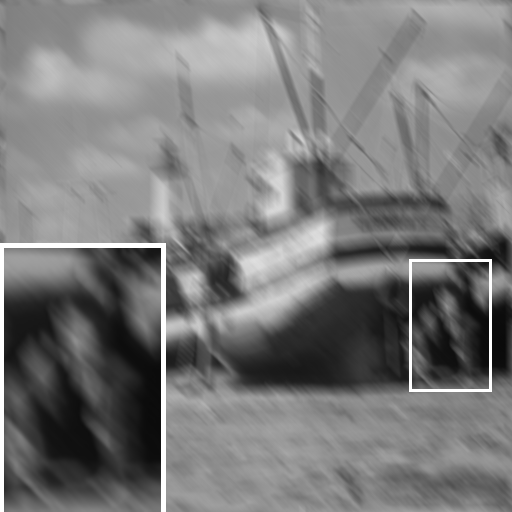}}
	}\\
	\makebox[0pt]{
		\subfigure[]{\label{fig:gull}\includegraphics[width=0.35\textwidth]{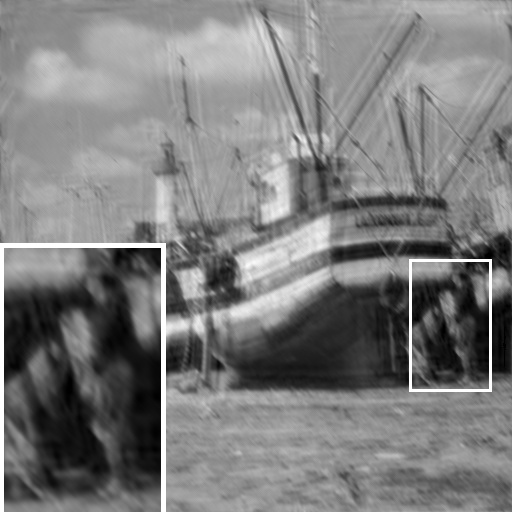}}
		\subfigure[]{\label{fig:tiger}\includegraphics[width=0.35\textwidth]{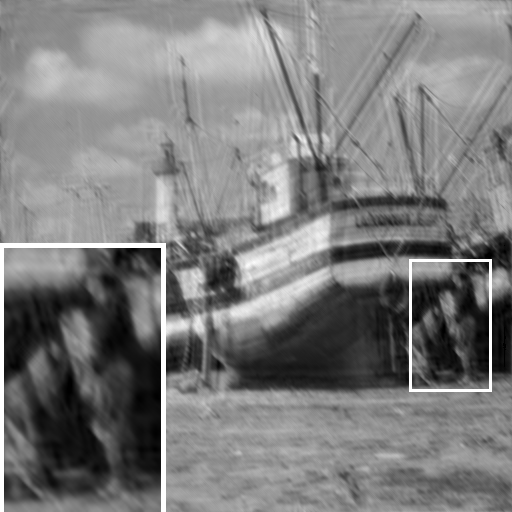}}
				\subfigure[FOTV \cite{17}]{\label{fig:tiger}\includegraphics[width=0.35\textwidth]{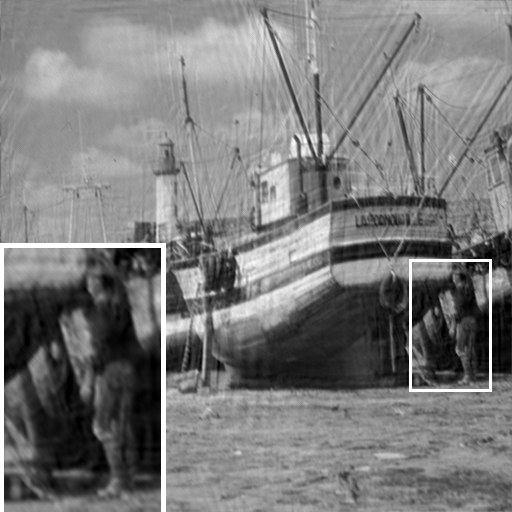}}
	}
	\vspace{-.2cm}
	\caption{(a) Ground-truth image, (b) degraded image with $k_4$, restored image by: 
		(c) method in \cite{2}, (d) method in \cite{1}, (e) proposed algorithm.}
	\label{fig6}
\end{figure}

\noindent As the last example, boat ($512 \times 512$) with 
$k_3=$ \texttt{fspecial($'$gaussian$'$,[19 19],2)} and
$k_4=$ \texttt{fspecial($'$motion$'$,20,135)} is studied.
In Table \ref{Tab3}, for different values of \texttt{std} 
and PSF, values PSNR, FSIM and SSIM are reported and these 
quantities are compared with methods in 
\cite{1,2}. It can be seen that the results of the proposed method for 
these values are better compared to the other methods.
Also, the results of the restored image are shown in Figure 
\ref{fig6}. In order to facilitate a more accurate comparison,
 a specific part of the image has been enlarged.
By comparing 
the results, it can be seen that the proposed method has better efficiency.

\begin{center}
	\begin{table}[H]
		\begin{threeparttable}
			\caption{Results for boat image with different values of \texttt{std} and PSF.}
			\label{Tab3}
			\centering
			\begin{tabular}{ccccccccccc}
				\hline
				&&\multicolumn{3}{c}{method in \cite{2}}   &\multicolumn{3}{c}{method in \cite{1}  }   &\multicolumn{3}{c}{proposed method}\\
				\cmidrule(l){3-5} \cmidrule(l){6-8} \cmidrule(l){9-11}
				PSF&\texttt{std} &PSNR      &FSIM   &SSIM  &PSNR         &FSIM&        SSIM      &PSNR      &FSIM     & SSIM\\
				\hline
        $k_3$	&0.005     &27.5232   &0.9320   &0.7438  &27.7284    &0.9458    &0.7660 &27.9938   &0.9568    &0.7768\\
				&0.0025    &27.6349   &0.9463  &0.7601 &27.9407     &0.9593   &0.7765   &29.0361    &0.9705    &0.8057\\
				\hline
		$k_4$	&0.005     &24.4325   &0.8523   &0.6723  &24.7457    &0.8628    &0.6843 &27.1650    &0.8971    &0.7527\\
		    	&0.0025    &24.9547   &0.8629  &0.6932 &25.1125     &0.8753   &0.7001   &28.0362    &0.9120    &0.7915\\	
			\hline
			\end{tabular}
		\end{threeparttable}
	\end{table}
\end{center}

\section{Conclusion} \label{sec5}

In this paper, semi-blind image deblurring is studied. Solving this type of problem is not easy because, unlike non-blind image deblurring, there is no complete information about the Point Spread Function (PSF). In the first step, a model to solve this problem using framelet transform and discrete fractional-order gradient is introduced. Then, a method based on ADMM is used to solve the proposed model. The proposed method improves the restored image by incorporating framelet and fractional calculations. The results of the proposed algorithm are compared with other methods, and these results confirm the effectiveness of our proposed algorithm.


\begin{thebibliography}{99}
%chicago
\providecommand{\doi}[1]{DOI~\discretionary{}{}{}#1}
%+++++++++++++++++++++++++++++++++++++++++++++++++
\bibitem{a5}
Adam, Tarmizi, and Raveendran Paramesran. "Hybrid non-convex 
second-order total variation with applications to non-blind image deblurring.
" Signal, Image and Video Processing 14, no. 1 (2020): 115-123.
\bibitem{b1}
Beck, Amir. First-order methods in optimization. Society for Industrial and Applied Mathematics, 2017.
\bibitem{1}
Dou, Hong-Xia, Ting-Zhu Huang, Xi-Le Zhao, Jie Huang, and Jun Liu. "Semi-blind image deblurring by a 
proximal alternating minimization method with convergence guarantees." Applied Mathematics and Computation 377 (2020): 125168.
\bibitem{a12}
Ganga, M., N. Janakiraman, Arun Kumar Sivaraman, Rajiv Vincent, A. Muralidhar, and Priya Ravindran. 
"An effective denoising and enhancement strategy for medical image using Rl-Gl-caputo method." 
Advances in Parallel Computing (Smart Intelligent Computing and Communication Technology) 38 (2021): 402-408.
\bibitem{a11}
Han, Bin. "Framelets and wavelets." Algorithms, Analysis, 
and Applications, Applied and Numerical Harmonic Analysis. Birkh{\"a}user xxxiii Cham (2017).
\bibitem{a3}
Hansen, Per Christian, James G. Nagy, and Dianne P. O'leary. Deblurring images: matrices, spectra,
 and filtering. Society for Industrial and Applied Mathematics, 2006.
\bibitem{a9}
He, Liangtian, Yilun Wang, and Zhaoyin Xiang. "Wavelet frame-based image restoration using sparsity, 
nonlocal, and support prior of frame coefficients." The Visual Computer 35, no. 2 (2019): 151-174.
\bibitem{a10}
Liu, Jing, Jieqing Tan, Xianyu Ge, Dandan Hu, and Lei He. "Blind deblurring with fractional-order 
calculus and local minimal pixel prior." Journal of Visual Communication and Image Representation 89 (2022): 103645.
\bibitem{a6}
Liu, Xinwu. "Total generalized variation and wavelet frame-based adaptive image restoration algorithm." 
The Visual Computer 35, no. 12 (2019): 1883-1894.
\bibitem{a8}
Ma, Liyan, Li Xu, and Tieyong Zeng. "Low rank prior and total variation regularization for 
image deblurring." Journal of Scientific Computing 70, no. 3 (2017): 1336-1357.
\bibitem{a7}
Pan, Jinshan, Deqing Sun, Hanspeter Pfister, and Ming-Hsuan Yang. "Blind image deblurring using 
dark channel prior." In Proceedings of the IEEE conference on computer vision and pattern 
recognition, pp. 1628-1636. 2016.
\bibitem{a1}
Parvaz, Reza. "Image restoration with impulse noise based on fractional-order total variation and 
framelet transform." Signal, Image and Video Processing (2023): 1-9.
\bibitem{a4}
Rudin, Leonid I., Stanley Osher, and Emad Fatemi. "Nonlinear total variation based noise removal 
algorithms." Physica D: nonlinear phenomena 60, no. 1-4 (1992): 259-268.
\bibitem{b2}
Sara, Umme, Morium Akter, and Mohammad Shorif Uddin. 
"Image quality assessment through FSIM, SSIM, MSE and PSNR - a comparative study.
" Journal of Computer and Communications 7, no. 3 (2019): 8-18.
\bibitem{a2}
Yin, M., Adam, T., Paramesran, R. and Hassan, M.F., 2022. An $l0$-overlapping group 
sparse total variation for impulse noise image restoration. Signal Processing: Image 
Communication, 102, p.116620.
\bibitem{2}
Zhao, Xi-Le, Wei Wang, Tie-Yong Zeng, Ting-Zhu Huang, and Michael K. Ng. "Total 
variation structured total least squares method for image restoration." SIAM Journal 
on Scientific Computing 35, no. 6 (2013): B1304-B1320.
\end{thebibliography}
\end{document}